\documentclass{article}
\usepackage[square,numbers]{natbib}
\usepackage{authblk}
\usepackage{charter}
\usepackage{fullpage}
\usepackage{bbm}

\usepackage{hyperref}       % hyperlinks
\usepackage{url}            % simple URL typesetting
\usepackage{booktabs}       % professional-quality tables
\usepackage{amsfonts}       % blackboard math symbols
\usepackage{nicefrac}       % compact symbols for 1/2, etc.
\usepackage{microtype}      % microtypography
\usepackage[table]{xcolor}         % colors
\usepackage{amsmath,amssymb,amsthm,amsfonts}
\usepackage{latexsym,bbm,graphicx,float,mathtools}

\usepackage[export]{adjustbox}
\usepackage{capt-of}% or \usepackage{caption}
\usepackage{booktabs}
\usepackage{parskip}

\usepackage[utf8]{inputenc}

\usepackage{wrapfig} 
\usepackage[titletoc]{appendix} % for \appendixpage

\usepackage{array}       
\usepackage[table]{xcolor} 
\usepackage{booktabs}   
\usepackage{multirow} 
\usepackage{longtable}
\usepackage{amsfonts} 
\usepackage{diagbox}      
\usepackage{titletoc}  % for 

\usepackage{microtype}
\usepackage{hyperref}
\usepackage{url}
\usepackage{booktabs}
\usepackage{amsmath}
\usepackage{amssymb}
\usepackage{mathtools}
\usepackage{amsthm}
\theoremstyle{plain}

\theoremstyle{definition}

\theoremstyle{remark}

\newcommand{\answerTODO}[1][]{\textcolor{red}{\bf [TODO]}}

\usepackage[noabbrev]{cleveref}
\usepackage{caption}
\usepackage{subcaption}
\usepackage{transparent}
\usepackage[inline]{enumitem}
\usepackage{multirow}
\usepackage{multicol}
\usepackage{algorithm}
\usepackage{algpseudocode}
\usepackage{xspace}

\usepackage{abstract}

\usepackage{tabularx}

\usepackage[subtle, mathdisplays=tight, charwidths=normal, leading=normal]{savetrees}

\title{\textbf{VADE: Variance-Aware Dynamic Sampling via Online Sample-Level Difficulty Estimation for Multimodal RL}}

\author{
    Zengjie Hu$^{1,2}$$^*$, Jiantao Qiu$^2$$^{\ddagger}$, Tianyi Bai$^{3}$$^*$, Haojin Yang$^1$, 
    \vspace{-0.6em} \\ 
    Binhang Yuan$^3$, Qi Jing$^1$$^{\dagger}$, Conghui He$^2$$^{\dagger}$, Wentao Zhang$^1$$^{\dagger}$
    \vspace{0.4em}\\
    $^1$Peking University,
    $^2$Shanghai AI Lab,
    $^3$HKUST
    \vspace{0.4em} \\
    ${}^*$ Equal contribution.\ ${}^\ddagger$ Project leader.\ ${}^\dagger$ Corresponding authors. \\
    {\{wentao.zhang, jingqi\}@pku.edu.cn, \{qiujiantao, heconghui\}@pjlab.org.cn}
}

\allowdisplaybreaks

\begin{document}
\setlength{\abovedisplayskip}{6pt}
\setlength{\belowdisplayskip}{6pt}
\setlength{\abovedisplayshortskip}{0pt}
\setlength{\belowdisplayshortskip}{3pt}

\maketitle

\begin{abstract}

Group-based policy optimization methods like GRPO and GSPO have become standard for training multimodal models, leveraging group-wise rollouts and relative advantage estimation. However, they suffer from a critical \emph{gradient vanishing} problem when all responses within a group receive identical rewards, causing advantage estimates to collapse and training signals to diminish. Existing attempts to mitigate this issue fall into two paradigms: filtering-based and sampling-based methods. Filtering-based methods first generate rollouts broadly and then retroactively filter out uninformative groups, leading to substantial computational overhead. Sampling-based methods proactively select effective samples before rollout but rely on static criteria or prior dataset knowledge, lacking real-time adaptability. To address these issues, we propose \textbf{VADE}, a \textbf{V}ariance-\textbf{A}ware \textbf{D}ynamic sampling framework via online sample-level difficulty \textbf{E}stimation. Our framework integrates three key components: online sample-level difficulty estimation using Beta distributions, a Thompson sampler that maximizes information gain through the estimated correctness probability, and a two-scale prior decay mechanism that maintains robust estimation under policy evolution. This  three components design enables VADE to dynamically select the most informative samples, thereby amplifying training signals while eliminating extra rollout costs. Extensive experiments on multimodal reasoning benchmarks show that VADE consistently outperforms strong baselines in both performance and sample efficiency, while achieving a dramatic reduction in computational overhead. More importantly, our framework can serves as a plug-and-play component to be seamlessly integrated into existing group-based RL algorithms. Code and models are available at https://VADE-RL.github.io.

\end{abstract}

\section{Introduction}
\label{sec:intro}

Recent advancements in Reinforcement Learning (RL) have markedly improved the performance of large language models (LLMs) and multimodal large language models (MLLMs) on complex tasks involving mathematical reasoning, code generation, and scientific problem-solving\cite{openai_o1, deepseek_r1, yang2025qwen3, vl_rethinker} . Reinforcement Learning with Verifiable Rewards (RLVR)\cite{RLVR} plays a central role in these developments and group-based policy optimization algorithms (e.g. GRPO\cite{grpo}, GSPO\cite{gspo}) have gained widespread adoption due to their efficiency and scalability. These methods generate multiple response rollouts for each training sample and compute relative advantages within the group for policy optimization, eliminating the need for additional value function approximation.

\begin{figure*}[htbp]
  \centering
   \includegraphics[width=\linewidth]{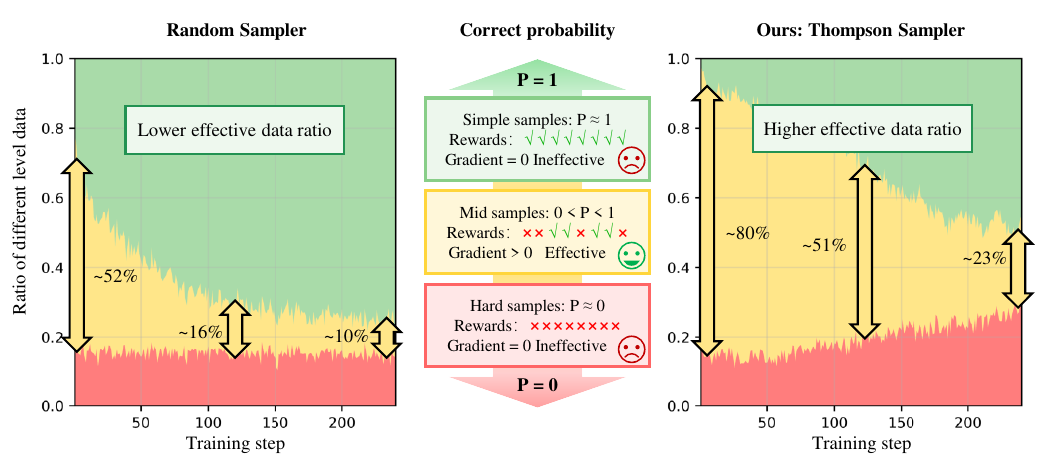}
   \caption{Proportion of data yielding effective training signals throughout training. \textbf{Left:} Under standard GRPO with random sampling, the proportion of informative samples providing non-zero gradients diminishes over time. \textbf{Right:} Our proposed Thompson sampling strategy consistently maintains a high ratio of effective data by proactively selecting informative samples.}
   \label{fig:correct_ratio}
\end{figure*}

%% 存在的问题定义  what is the problem?
Despite their success, group-based RL methods suffer from a critical limitation known as the \emph{gradient vanishing} problem\cite{advantage_vanishing}: when all responses within a group receive identical rewards—a common scenario when the model either solves all problems correctly or fails entirely—the relative advantages collapse to zero. This effectively eliminates the training signal for the entire group, severely degrading sample efficiency. As illustrated in the left half of \Cref{fig:correct_ratio}, during standard GRPO training with random sampling, the proportion of data yielding effective gradients progressively decreases, with zero-gradient groups dominating the training batches. In later stages, only around 10\% of samples provide useful learning signals, creating a fundamental bottleneck for efficient training.

%% 解决这个问题的困难之处 Why is it hard?
% Addressing this problem presents a two-fold challenge in efficiently identifying informative samples. First, for groups where the model consistently fails, generating more rollouts is a computationally prohibitive solution for large multimodal models; for groups that are totally solved, additional rollouts are futile, and the key is to avoid wasting resources on them. Second, the learning process is highly non-stationary; as the policy model evolves, the difficulty of given samples changes dynamically, necessitating a real-time and adaptive selection strategy.

%% 现有方法及其问题  Why hasn't it been solved before?

Existing attempts to mitigate this issue can be categorized into two paradigms: filtering-based and sampling-based methods, both of which exhibit critical limitations. Filtering-based methods, such as DAPO\cite{dapo} and VCRL\cite{vcrl}, first generate rollouts for a broad set of samples and then retroactively filter out groups with uninformative reward signals. This "generate-then-filter" approach inevitably leads to substantial computational waste and requires costly mechanisms like repeated rollouts or replay buffers to assemble valid batches. In contrast, sampling-based methods, such as SEC\cite{sec} and MMR1\cite{mmr1}, aim to proactively select samples of appropriate difficulty before rollout generation. However, they rely on static criteria or prior dataset knowledge: SEC depends on pre-defined, coarse-grained difficulty categories, while MMR1 requires periodic and expensive re-rollouts to refresh its sample-level estimates. Consequently, both strategies lack the fine-grained, real-time adaptability necessary to keep pace with an evolving policy, leaving an open need for a dynamic and efficient sample selection mechanism.

%% 我们的解决方案和效果

To address these limitations, we propose VADE, a Variance-Aware Dynamic sampling framework via online difficulty Estimation, designed to overcome the gradient vanishing problem in group-based RL. VADE introduces an efficient, sample-level data selection mechanism that operates without additional rollout inference overhead. We formulate data selection as a non-stationary Multi-Armed Bandit problem and solve it with a dynamic Thompson sampling strategy that continuously adapts to the policy's evolution.  Our framework integrates three key components: an online sample-level difficulty estimation module that employs a Beta distribution as the posterior estimate of each sample's correctness probability; a Thompson sampler which leverages these estimates to maximize information gain and balance exploration with exploitation; and a two-scale prior decay mechanism that robustly maintains estimation accuracy under policy evolution. This integrated approach enables VADE to proactively identify the most informative samples, thereby maximizing the proportion of data yielding effective gradients, as visualized in the right half of \Cref{fig:correct_ratio}. Notably, VADE requires no prior difficulty annotations, incurs no extra rollout cost, and serves as a plug-and-play component that can be seamlessly integrated into existing group-based RL algorithms such as GRPO and GSPO. Comprehensive experiments on multimodal reasoning benchmarks across various model scales and policy optimization algorithms show that VADE consistently achieves higher sample efficiency and better evaluation performance compared to strong baselines, demonstrating its robustness and generalization. Furthermore, ablation studies validate the effectiveness of each component in our framework.

%% 现在main contribution的思路
%%% 第一点突出我们提出了解决gradient vanishing problem的framework，（强调工作流程和解决实际问题的能力）
%% 第二点强调formalization，问题的建模和解决方法，相当于理论性的贡献
%% 第三点是实验效果的证明

Our main contributions are summarized as follows:
\begin{itemize}
    % \item We propose VADE, a novel and efficient framework designed to tackle the gradient vanishing problem in group-based RL. It integrates three key components: an online sample-level difficulty estimation using Beta distributions, a Thompson Sampler for dynamic data selection, and a two-scale prior decay mechanism to update and maintain robust estimates.
    \item We propose VADE, a novel and efficient framework designed to tackle the gradient vanishing problem in group-based RL. It enables dynamic selection of highly informative samples during training without introducing extra rollouts, thereby achieving higher performance and training efficiency.
    \item We establish a theoretical formulation by modeling data selection as a non-stationary Multi-Armed Bandit problem and then solve it using a dynamic Thompson Sampling strategy. This formulation requires no prior difficulty knowledge or annotated categories and can serve as a plug-and-play component for existing group-based RL algorithms.
    \item We conduct systematic experiments on multimodal reasoning tasks across various model scales and policy optimization algorithms. The experimental results show that VADE consistently achieves higher sample efficiency and better evaluation performance compared to strong baselines, demonstrating its robustness and generalization.
\end{itemize}

\section{Related Work}
\label{sec:related_work}

The \emph{gradient vanishing} problem presents a fundamental bottleneck for group-based policy optimization algorithms. Existing approaches to address this issue can be categorized into two main paradigms based on when and how they select informative data.\\
\textbf{Filtering-based Methods.} This line of work operates by generating rollouts first and filtering out non-informative samples afterward\cite{dapo, vcrl,e2h, vl_cogito}. DAPO \cite{dapo} exemplifies this approach by discarding groups with zero reward variance and performing repeated rollouts to assemble valid batches. VCRL \cite{vcrl} extends this idea by using reward variance as a difficulty proxy, filtering out samples with variance below a pre-defined threshold and replenishing the batch from a replay buffer. While effective in ensuring gradient quality, these methods inherently follow a "generate-then-filter" paradigm that leads to substantial computational waste due to discarded rollouts. In contrast, our VADE framework avoids this overhead by proactively selecting samples before the rollout stage, eliminating the need for costly filtering and re-rollout mechanisms.\\
\textbf{Sampling-based Methods.} In contrast, this category aims to select high-utility samples \emph{before} rollout generation\cite{sec, mmr1, knapsackrl}. Curriculum learning methods like SEC \cite{sec} formulate sample selection as a Multi-Armed Bandit problem over pre-defined difficulty categories. Meanwhile, MMR1 \cite{mmr1} employs a variance-aware sampling strategy that combines reward variance and trajectory diversity. However, these approaches rely on static criteria: SEC depends on coarse-grained, pre-defined difficulty hierarchies, while MMR1 requires periodic and expensive re-rollouts to refresh its estimates. Consequently, they lack the fine-grained, real-time adaptability needed to keep pace with an evolving policy. Unlike these methods, VADE performs fully online, sample-level difficulty estimation without relying on pre-defined categories or expensive re-rollouts, enabling real-time adaptation to policy evolution through its novel two-scale prior decay mechanism.

\begin{figure*}[t]
  \centering
   \includegraphics[width=\linewidth]{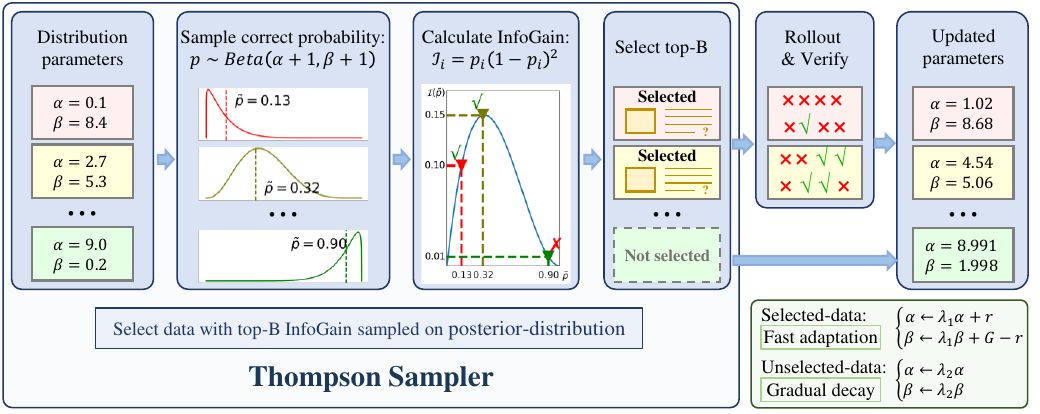}
   \caption{\textbf{Overview of the VADE framework.} Our method maintains distributions $\text{Beta}(\alpha_i + 1, \beta_i+ 1)$ for each sample to enable online difficulty estimation. Through Thompson sampling and InfoGain $\mathcal{I}_i = p_t(1-p_t)^2$ maximization, VADE dynamically selects informative batches for group-wise rollouts. The two-scale prior decay mechanism ensures estimates remain accurate throughout policy evolution.}
   \label{fig:main_framework}
\end{figure*}

\section{Method}

In this section, we present the VADE framework, which tackles the \emph{gradient vanishing} problem by selecting the most informative samples to maximize learning signals. An overview of the framework is illustrated in \Cref{fig:main_framework}. This section is organized as follows: In \Cref{subsec:info_gain}, we introduce the background of group-based policy optimization and formalize the sample selection objective from an information gain perspective. In \Cref{subsec:thompson_sampling}, we formulate the data selection process as a non-stationary Multi-Armed Bandit problem and details its solution via dynamic Thompson sampling strategy. Finally, in \Cref{subsec:online_estimation}, we introduce the two-scale prior decay mechanism for maintaining robust online sample-level estimates under policy evolution. The complete RL training pipeline of our VADE framework is illustrated in \Cref{algo:our_method}.

\subsection{Group-wise Policy Optimization and Information Gain Theory}
\label{subsec:info_gain}
Current mainstream RLVR methods for training vision-language models follow this setup: each training data point \(x_i\) consists of an image and a related question, along with a reward function \(R(x_i, y_i) \rightarrow \{0,1\}\) that verifies the answer and assigns a reward score. For a question \(x_i\) and its corresponding answer \(y_i\), \(R(x_i, y_i) = 1\) indicates that the verification function considers the answer correct, while \(R(x_i, y_i) = 0\) indicates otherwise.

For a policy function \(\pi_{\theta}(y|x)\) with parameters \(\theta\), the optimization objective on the training set \(\mathcal{D}\) is:
\begin{equation}
    J(\theta) = \mathbb{E}_{x \sim \mathcal{D}} \mathbb{E}_{y \sim \pi_{\theta}(\cdot|x)} [R(x, y)]
\end{equation}

The most popular group-wise policy optimization methods, such as GRPO and GSPO, employ group-wise rollouts to sample data and compute advantages based on the reward distribution within each rollout group:
\begin{equation}
    A_{i, j} = \frac{R_{i, j} - mean(\{R_{i, j}\}_{j=1}^G)}{std(\{R_{i, j}\}_{j=1}^G)}
\end{equation}
This approach eliminates the need for an additional value network, enabling efficient training with a reliable reward baseline.

A widely observed phenomenon is that the performance gain \(\mathcal{I}_i\) from training on a specific data point \(x_i\) is related to its current difficulty\cite{knapsackrl}. According to the theory of sample information gain in group-wise policy optimization methods, as presented in \textit{Knapsack RL}\cite{knapsackrl}, for a problem \(x_i\), let \(p_i(\theta)\) denote the probability of correct reasoning under the current policy model \(\pi_{\theta}\):
\begin{equation}
    p_i(\theta) = P_{y\sim \pi_{\theta}(\cdot|x_i)} (R(x_i, y)=1)
\end{equation}
The information gain \(\mathcal{I}_i(\theta)\) from selecting this data point for policy gradient optimization is approximately:
\begin{equation}
    \mathcal{I}_i(\theta) \approx p_i(\theta)(1 - p_i(\theta))^2
    \label{eq:info_gain}
\end{equation}
Under the constraints of a fixed batch size \(B\) per training step and a rollout budget \(G\) per \(x_i\), selecting the top \(B\) data points with the highest \(\mathcal{I}_i(\theta)\) theoretically maximizes the single-step training gain.

\subsection{Non-stationary Multi-Armed Bandit with Thompson Sampling}
\label{subsec:thompson_sampling}
However, in practice, directly computing \(p_i(\theta)\) precisely is infeasible due to the complexity of neural networks. To address this, we characterize its distribution \(f(p_i(\theta))\) using rollout results and reward verification during training. Given that group-wise policy optimization inherently relies on rollout results from the current training batch, we naturally model the online data sampling and training process as a \textit{Non-stationary Multi-Armed Bandit}:

\begin{enumerate}
    \item At training step \(t\), estimate the distribution of correctness probabilities \(f(p_i(\theta_t) | h_{i,1:t})\) for all data points based on historical rollout records \(h_{i,1:t}\), where \(h_{i,1:t}\) represents the rollout history of \(x_i\).
    \item Compute \(\mathcal{I}_i(\theta_t)\) using \(f(p_i(\theta_t) | h_{i,1:t})\) and sample a batch of training data \(\mathcal{B}_t = [x_{1,t}, x_{2,t}, \cdots, x_{B,t}]\). Perform rollouts on these data points and update the model parameters.
    \item Update the rollout results for the current training step into \(h_{i,1:t+1}\).
\end{enumerate}

For a data point \(x_i\) at training step \(t\), two scenarios are possible:  
a) It is sampled and generates \(G\) trajectories, with \(a_{i,t}\) correct responses;  
b) It is not sampled, denoted as \(\emptyset\).  
Thus, the historical record \(h_{i,1:t}\) can be represented as a sequence containing \(a_{i,t}\) and \(\emptyset\), e.g., \([a_{i,1}, \emptyset, a_{i,3}, \cdots]\).

Once the posterior distribution \(f(p_i(\theta_t) | h_{i,1:t})\) is estimated, sampling data according to the Thompson Sampling\cite{thompson_sampling} method theoretically achieves high sample efficiency. Specifically, we sample a candidate probability \(\tilde{p}_i(\theta_t)\) from the posterior distribution \(f(p_i(\theta_t) | h_{i,1:t})\) and compute the corresponding information gain \(\tilde{\mathcal{I}}_i(\theta_t)\). We then select the top \(B\) data points with the highest \(\tilde{\mathcal{I}}_i(\theta_t)\) to form the current training batch. This strategy naturally balances exploration and exploitation: while the information gain objective \(\tilde{\mathcal{I}}_i(\theta_t)\) favors samples with high expected utility, the random sampling from the posterior distribution  \(f(p_i(\theta_t) | h_{i,1:t})\) gives those uncertain yet potentially valuable samples a chance of being selected.

% \subsection{Online Estimation Distribution of Correct Probability}
\subsection{Online Estimation with Two-Scale Prior Decay}
\label{subsec:online_estimation}
To implement the Thompson sampling strategy described above, we need a probabilistic model for the correctness probability $p_i$. A natural choice is the Beta distribution\cite{beta_distribution}, which is a conjugate prior for the Bernoulli process. We maintain parameters \((\alpha_i, \beta_i)\) for each data point $x_i$, where $\alpha_i$ and $\beta_i$ represent the cumulative counts of correct and incorrect responses for $x_i$, respectively, and we model the correctness probability with distribution $\text{Beta}(\alpha_i + 1, \beta_i + 1)$ to ensure numerical stability. After rolling out \(G\) responses and observing $r_i$ correct and $G - r_i$ incorrect answers, the posterior distribution becomes \(\text{Beta}(\alpha_i + 1 + r_i, \beta_i + 1 + G - r_i)\) under a stationary environment.

However, the model's continuous evolution during training makes the correctness probability non-stationary. If we simply updated the Beta parameters by adding new outcomes (e.g. $\alpha_i = \alpha_i + r_i$), the historical data would eventually overwhelm the new evidence, leading to outdated estimates. This is particularly problematic in two scenarios: (1) When a sample is used for training, its response distribution shifts significantly, diminishing the value of past rollouts; (2) Conversely, even unused samples are affected by generalization from trained data, gradually making their historical estimates less relevant.

Based on this analysis, we propose a \textbf{\textit{two-scale prior decay model}} with two decay parameters \(0 < \lambda_1 \ll \lambda_2 < 1\) to account for the two scenarios:

\begin{itemize}
    \item If $x_i$ is sampled:
    \begin{equation}
        \left\{
            \begin{aligned}
                &\alpha_{i,t} \leftarrow \lambda_1 \alpha_{i,t-1} + r_t\\ 
                &\beta_{i,t} \leftarrow \lambda_1 \beta_{i,t-1} + G - r_t
            \end{aligned}
        \right.
        \label{eq:our_update_a_b_sampled}
    \end{equation}
    
    \item If $x_i$ is not sampled:
    \begin{equation}
        % \end{aligned}
        \left\{
            \begin{aligned}
                &\alpha_{i,t} \leftarrow \lambda_2 \alpha_{i,t-1}\\ 
                &\beta_{i,t} \leftarrow \lambda_2 \beta_{i,t-1}
            \end{aligned}
        \right.
        \label{eq:our_update_a_b_not_sampled}
    \end{equation}
\end{itemize}
This design accounts for: (1) significant distribution shift when data is directly used for training (\(\lambda_1\)), and (2) gradual generalization effects when data is not selected (\(\lambda_2\)).

\Cref{algo:our_method} provides the complete training pipeline for our VADE framework.

\begin{algorithm}
    \caption{Reinforcement Learning with VADE.}
    \label{algo:our_method}
    \begin{algorithmic}[1]
        \Require Dataset $\mathcal{D}$, batch size $B$, rollouts per sample $G$, decay rates $\lambda_1, \lambda_2$, training steps T, policy model $\pi_{\theta}$
        \For{each sample $x_i \in \mathcal{D}$}
            \State Generate G rollouts $\{y_i\}_{i=1}^G$ from $\pi_{\theta}(\cdot | x)$
            \State Initial $\alpha_i, \beta_i$ according to the rollouts results.
        \EndFor

        \For{training step t=1, $\dots$, T}
            \For{each sample $x_i$ $\in \mathcal{D}$}
                \State Compute $p_i$ using Thompson sampling. 
                \State Compute $\mathcal{I}_i(\theta)$ according \cref{eq:info_gain}
            \EndFor
            % \State Select the top-$B$ samples with the highest $\mathcal{I}_i(\theta)$ to form the training batch $\mathcal{B}$.
            \State Select the top-$B$ $\mathcal{I}_i(\theta)$ to form the training batch $\mathcal{B}$.
            
            \State Apply RL update using the Training Batch $\mathcal{B}$
            \For{each sample $x_i$ $\in \mathcal{D}$}
                \State Update $\alpha_i, \beta_i$ according to \cref{eq:our_update_a_b_sampled} and \cref{eq:our_update_a_b_not_sampled}.
            \EndFor
        \EndFor
    \end{algorithmic}
\end{algorithm}

% \cellcolor{red!5}
% \cellcolor{green!10}

\begin{table*}[!htbp]
  \centering
  \caption{Main performance comparison of VADE against other RL baselines on Qwen2.5-VL models. Highlighted cells show whether using the method \colorbox{green!10}{improves} or \colorbox{red!5}{degrades} the performance compared to the vanilla method.}
  \label{table:main_results}
  \begin{tabular}{cccccccc}
    \toprule
    \multirow{2}{*}{Algo.} & \multirow{2}{*}{Method}& \multicolumn{5}{c}{Average@4} & \multirow{2}{*}{AVG} \\
    \cmidrule{3-7}
    & & MathVista & MathVerse & MathVision & ScienceQA & ChartQA & \\
    \midrule
    \rowcolor{gray!10}
    \multicolumn{8}{c}{Qwen2.5-VL-7B-Instruct} \\
    \midrule
    \multicolumn{2}{c}{Base Model} & 69.30 & 40.36  & 18.55 & 84.53 & 80.40 & 58.63 \\
    \midrule
    \multirow{4}{*}{GRPO} 
      & vanilla & 72.85 & 41.36  & 20.71 & 90.08 & 84.12 & 61.82 \\
      & DAPO & \cellcolor{red!5}71.80 & \cellcolor{green!10}\textbf{47.11}  & \cellcolor{green!10}24.16 & \cellcolor{red!5}89.54 & \cellcolor{green!10}\textbf{85.72} & \cellcolor{green!10}63.67 \\
      % & MMR1  & --    &  --    &  --   &  --   &  --   &      -- \\
      & \textbf{Ours} & \cellcolor{green!10}\textbf{75.10} & \cellcolor{green!10}45.65  & \cellcolor{green!10}\textbf{25.79} & \cellcolor{green!10}91.67 & \cellcolor{green!10}85.56 & \cellcolor{green!10} \textbf{64.75} \\
    \midrule
    \multirow{3}{*}{GSPO}
      & vanilla & 71.88 & 41.09  & 23.29 & \textbf{91.92} & 84.60 & 62.56 \\
      & DAPO & \cellcolor{green!10}73.35 & \cellcolor{green!10}\textbf{47.01}  & \cellcolor{green!10}24.13 & \cellcolor{red!5}91.32 & \cellcolor{green!10}86.04 & \cellcolor{green!10}64.37 \\
      & \textbf{Ours} & \cellcolor{green!10}\textbf{73.45} & \cellcolor{green!10}45.70  & \cellcolor{green!10}\textbf{25.68} & \cellcolor{red!5}91.77 & \cellcolor{green!10}\textbf{86.24} & \cellcolor{green!10}\textbf{64.57} \\
    \midrule
    \rowcolor{gray!10}
    \multicolumn{8}{c}{Qwen2.5-VL-3B-Instruct} \\
    \midrule
    \multicolumn{2}{c}{Base Model} & 63.15 & 23.27 & 11.29 & 76.70 & 78.84 & 50.65 \\
    \midrule
    \multirow{4}{*}{GRPO}
      & vanilla & 63.37 & 33.40 & 19.16 & 85.08 & 76.80 & 55.56 \\
      & DAPO & \cellcolor{green!10}\textbf{66.88} & \cellcolor{green!10}\textbf{38.03} & \cellcolor{green!10}\textbf{21.82} & \cellcolor{red!5}83.64 & \cellcolor{green!10}\textbf{82.20} & \cellcolor{green!10}\textbf{58.51} \\
      % & MMR1 & -- & -- & -- & -- & -- & -- \\
      & \textbf{Ours}& \cellcolor{green!10}63.62 & \cellcolor{green!10}37.22 & \cellcolor{red!5}18.92 & \cellcolor{green!10}\textbf{85.37} & \cellcolor{green!10}81.48 & \cellcolor{green!10}57.32 \\
    \midrule
    \multirow{3}{*}{GSPO}
      & vanilla  & 62.80 & \textbf{33.47} & 17.55 & 84.09 & \textbf{82.12} & 56.01 \\
      & DAPO & \cellcolor{green!10}\textbf{64.70} & \cellcolor{red!5}32.08 & \cellcolor{red!5}16.00 & \cellcolor{red!5}81.36 & \cellcolor{red!5}81.36 & \cellcolor{red!5}55.10 \\
      & \textbf{Ours} & \cellcolor{green!10}63.82 & \cellcolor{red!5}32.89 & \cellcolor{green!10}\textbf{19.20} & \cellcolor{green!10}\textbf{84.58} & \cellcolor{red!5}81.40 & \cellcolor{green!10}\textbf{56.38} \\
    \bottomrule
  \end{tabular}
\end{table*}

\section{Experiments}
We conduct comprehensive experiments to evaluate the effectiveness and efficiency of our proposed VADE framework across a range of visual reasoning tasks. This section is organized as follows: In \Cref{subsec:experimental_setup}, we describe our experimental setup, including model configurations, training details, and evaluation benchmarks. \Cref{subsec:main_results} presents the main results comparing our method against strong baselines across multiple visual reasoning benchmarks. \Cref{subsec:training_dynamic} presents the training dynamics to demonstrate the efficiency of our approach during the training process. Finally, in \Cref{subsec:ablation_study}, we provide ablation studies to validate the design choices of our method.

\subsection{Experimental Setup}
\label{subsec:experimental_setup}
\textbf{Implementation Details:} 
We conduct reinforcement learning training on both Qwen2.5-VL-7B-Instruct and Qwen2.5-VL-3B-Instruct models \cite{qwen25vl}, implementing all experiments using the verl \cite{verl} RL training framework. We implement our proposed VADE method using both GRPO \cite{grpo} and GSPO \cite{gspo} algorithms, with comparative analysis against vanilla GRPO/GSPO and DAPO \cite{dapo} as baselines. It should be noted that, to ensure a fair comparison, we only utilize the dynamic sampling component of DAPO without incorporating its other configurations. Therefore, we denote this implementation as DAPO + GRPO/GSPO. The experimental configuration is standardized across all methods: training batch size is set to 512 and we sample 8 rollouts per prompt. The epoch of training is 15. In the implementation of our VADE method, the update parameters are configured as $\lambda_1=0.2$ and $\lambda_2=0.999$. For vanilla GRPO/GSPO and our proposed method, we complete all scheduled training steps. For DAPO, due to its need to over-sample from the training data, we limit the maximum number of rollout generation to three times the total training steps to ensure fair comparison. For further implementation details and experimental results, please refer to Appendix. \\% \Cref{app:experiment_details} For MMR1, which is officially implemented only with the GRPO algorithm, we use the code provided by the original authors but align its parameter settings with ours, while retaining their strategy of updating weight estimates every $T_{\text{update}} = 45$ steps as described in their paper.
\textbf{Dataset:} For the training dataset, we select 9 tasks from LLaVA-OneVision-Data\cite{llavaov} and incorporate the training sets of ChartQA\cite{masry2022chartqa} and ScienceQA\cite{ScienceQA}, ensuring no overlap between the training data and the evaluation benchmarks. The constructed dataset spans three distinct domains: mathematics, chart understanding, and scientific reasoning, comprising a total of 9,471 samples. During training, we split the dataset into 90\% for training and 10\% for validation. Further details regarding the dataset are provided in Appendix. To specifically demonstrate the effectiveness of our proposed method, all models and approaches in our experiments are exclusively trained using this dataset through Reinforcement Learning.\\
\textbf{Evaluation Benchmarks:} 
We conduct comprehensive evaluation across five challenging benchmarks: MathVista \cite{lu2023mathvista}, MathVerse \cite{zhang2024mathverse} and MathVision \cite{MathVision} for mathematical reasoning, ScienceQA \cite{ScienceQA} for scientific question answering, and ChartQA \cite{masry2022chartqa} for chart comprehension. For the MathVista, MathVerse and MathVision benchmarks, we employ the official evaluation code provided in their original papers, which utilizes the \texttt{LLM-as-a-judge} approach for answer assessment. In our implementation, we use the Qwen2.5-72B-Instruct\cite{qwen_2_5} model deployed with vLLM\cite{vllm} as the judge model. For remaining benchmarks, we utilize the lmms-eval \cite{lmms-eval} framework for evaluation. To ensure stable and reliable evaluation results, we report the average performance across 4 independent runs (average@4) for all experiments.

\begin{figure*}[htbp]
   \centering
   \includegraphics[width=\linewidth]{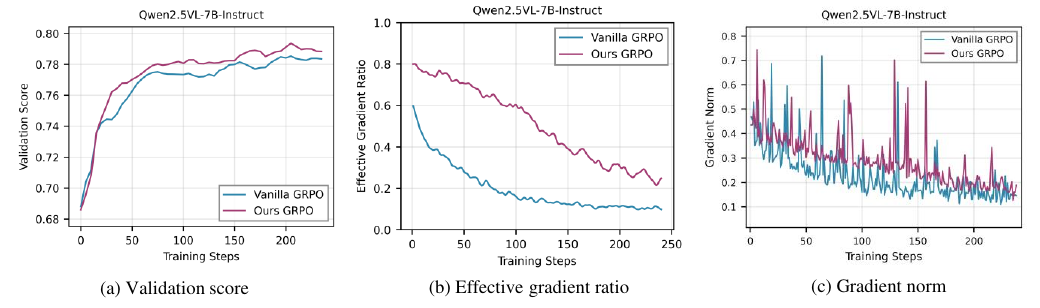}

   \caption{Training dynamics of Qwen2.5VL-7B-Instruct trained with the GRPO algorithm. (a) \textbf{Validation score}, illustrating convergence speed and final performance; (b) \textbf{Effective gradient ratio}, representing the proportion of data with non-uniform rewards (neither all-zero nor all-one) in each training batch, reflecting data efficiency; (c) \textbf{Actor gradient norm}, indicating the magnitude of gradient signals throughout training.}
   \label{fig:training_dynamic_ours_vanilla}
\end{figure*}

\subsection{Main Results}
\label{subsec:main_results}

\begin{figure*}
  \centering
  \includegraphics[width=\linewidth]{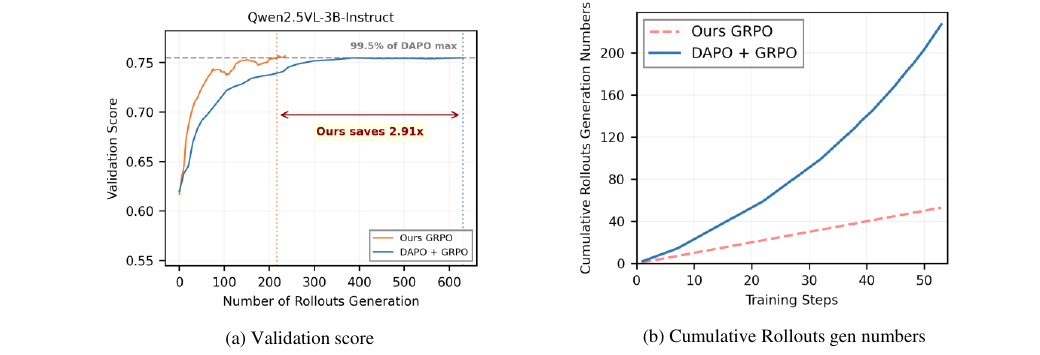} 
  \caption{Training dynamics comparison between our method and DAPO. (a) \textbf{Validation Score.} The x-axis represents the total count of forward passes through the model to sample responses for all training batches. Our method achieves competitive or superior validation performance with significantly fewer rollout generations, demonstrating substantially higher training efficiency compared to DAPO. (b)\textbf{Cumulative Rollout Generations throughout training.} This figure plots the total number of forward passes performed to sample responses for all training batches up to a given step. DAPO incurs significantly more rollout generations than our VADE due to its over-sampling and filtering strategy, demonstrating the superior computational efficiency of our approach.}
  \label{fig:training_dynamic_dapo}
\end{figure*}

We evaluate ours VADE method on multimodal reasoning benchmarks by training Qwen2.5-VL-3B/7B-Instruct models on our constructed dataset and comparing against vanilla GRPO/GSPO and DAPO baselines. As shown in \Cref{table:main_results}, ours VADE consistently outperforms baseline methods, demonstrating its both effectiveness and efficiency.\\
\textbf{Our approach demonstrates stable improvements over vanilla methods.} After training, our method achieves higher average scores than vanilla baselines across all configurations. Specifically, with Qwen2.5VL-7B-Instruct, our method outperforms vanilla GRPO by 2.93\% and vanilla GSPO by 2.01\% in average performance. Across 20 specific measurements (5 benchmarks × 4 configurations), our method surpasses vanilla approaches in 16 cases, with minimal performance gaps in the remaining 4 cases. These results consistently validate the effectiveness of our sample selection strategy.\\
\textbf{Our method achieves better performance than DAPO with significantly less computational overhead.} While DAPO relies on an "over-sample and filter" paradigm that incurs growing computational overhead, VADE achieves better or comparable performance with only one-third of DAPO's rollout budget. Under this constrained setting, VADE achieves higher average scores in 3 out of 4 configurations. More importantly, as shown in \Cref{fig:training_dynamic_dapo}(a), our method achieves higher validation scores with far fewer rollout generations. Remarkably, VADE attains 99.5\% of DAPO's peak performance while using nearly three times fewer rollout inferences, highlighting its superior sampling efficiency and practical scalability. \\
\textbf{Our performance gains stem from selecting more informative training data.} VADE achieves consistent performance gains by dynamically estimating sample difficulty and selecting the most informative data at each training step, thereby enhancing the learning signal and leading to more effective policy updates. As evidenced in \Cref{fig:correct_ratio} and \Cref{fig:training_dynamic_ours_vanilla}(b), our method maintains a significantly higher proportion of data yielding effective gradients by consistently selecting informative samples. Moreover, unlike DAPO, which wastes computation on repeated rollouts for filtering, VADE leverages each rollout result to update online difficulty estimates, avoiding extra inference costs. This design enables more efficient parameter updates and faster convergence compared to both vanilla optimization algorithms and filtering-based method.\\

% \begin{figure}[t]
%    \centering
%    \includegraphics[width=\linewidth]{figure/score_dapo_ours/3b_grpo_moving_average.png} 
%    \caption{Training dynamics comparison between our method and DAPO. The x-axis represents the total count of forward passes through the model to sample responses for all training batches. Our method achieves competitive or superior validation performance with significantly fewer rollout generations, demonstrating substantially higher training efficiency compared to DAPO.}
%    \label{fig:training_dynamic_ours_dapo}
% \end{figure}

% \begin{figure}[t]
%    \centering
%    \includegraphics[width=\linewidth]{figure/dapo_cumulative_gen_steps.png} 
%    \caption{\textbf{Cumulative Rollout Generations throughout training.} This figure plots the total number of forward passes performed to sample responses for all training batches up to a given step. DAPO incurs significantly more rollout generations than our VADE due to its over-sampling and filtering strategy, demonstrating the superior computational efficiency of our approach.}
%    \label{fig:dapo_gen_nums}
% \end{figure}

\subsection{Training Dynamics}
\label{subsec:training_dynamic}
To further validate the efficiency and effectiveness of our VADE, we analyze key training dynamics against baseline methods. \Cref{fig:training_dynamic_ours_vanilla} illustrates the training dynamics of Qwen2.5VL-7B-Instruct using GRPO algorithm. Additional results for other model configurations are provided in the Appendix. We focus on three critical aspects: validation score, effective gradient ratio, and actor gradient norm:
\begin{itemize}
    \item The \textbf{validation score} serves as the primary indicator of model capability during training. As shown in \Cref{fig:training_dynamic_ours_vanilla}(a), VADE not only achieves higher final performance but also demonstrates accelerated convergence compared to vanilla GRPO, coonfirming its ability to improve final model performance.
    \item The \textbf{effective gradient ratio} measures the proportion of samples that yeilds effective gradients in each training batch.  \Cref{fig:training_dynamic_ours_vanilla}(b) reveals a striking difference: while vanilla GRPO suffers from rapidly diminishing effective gradients, VADE maintains a consistently high ratio throughout training. This fundamental improvement directly addresses the core \emph{gradient vanishing} problem by ensuring that most training batches contain diverse reward outcomes. The sustained high ratio demonstrates VADE's ability to continuously identify and select samples that challenge the current policy without being trivially easy or impossibly difficult.
    \item The \textbf{actor gradient norm} reflects the magnitude of parameter updates to the policy model during training. As shown in \Cref{fig:training_dynamic_ours_vanilla}(c), VADE produces significantly higher gradient norms compared to vanilla GRPO, indicating more substantial and directed policy updates. This enhancement is achieved through strategic prioritization of samples that provide informative gradient signals, effectively amplifying the learning signal per parameter update.
    
\end{itemize}

Furthermore, we compare the computational efficiency of VADE against DAPO. As DAPO relies on repeated rollouts to filter uninformative groups through its "generate-then-filter" paradigm, we use the total number of rollout generations as a fair efficiency metric. \Cref{fig:training_dynamic_dapo}(b) clearly shows that DAPO incurs substantially more rollout inferences due to this repetitive sampling strategy. The efficiency gap stems from their fundamental difference: DAPO wastes computation on generating and discarding rollouts, while VADE's proactive selection ensures most rollouts contribute to learning. Remarkably, as shown in \Cref{fig:training_dynamic_dapo}(a), VADE achieves 99.5\% of DAPO's peak performance while using several times fewer inference computations. This demonstrates that VADE provides a highly inference-efficient alternative, offering strong performance without the computational burden of oversampling.

\subsection{Ablation Study}
\label{subsec:ablation_study}
% We conduct ablation studies to validate the design of VADE's three core components: the online difficulty estimator based on Beta distributions and an information gain objective, the Thompson sampler for stochastic data selection, and the two-scale prior decay mechanism for robust parameter updates. \Cref{tab:ablation_study} compares variants of our method when training Qwen2.5VL-7B-Instruct with GRPO.\\
We conduct ablation studies to validate the design of VADE's three core components: the online difficulty estimator based on information gain objective, the Thompson sampler for balancing exploration and exploitation, and the two-scale prior decay mechanism for robust parameter updates. \Cref{tab:ablation_study} compares variants of our method when training Qwen2.5VL-7B-Instruct with GRPO, reporting the average performance across our five evaluation benchmarks.\\
\textbf{Our asymmetric information gain $p(1-p)^2$ outperforms symmetric formulation $p(1-p)$.} We first evaluate the information gain objective that guides the online estimator, comparing our asymmetric formulation $\mathcal{I}_i = p(1-p)^2$ against the symmetric alternative $p(1-p)$ used in MMR1\cite{mmr1}. The former actively prioritizes challenging samples, while the latter treats all non-extreme cases uniformly. Results show that our formulation improves average performance by 1.09\%, confirming that emphasizing harder problems is beneficial for model advancement.\\
\textbf{Our Thompson sampling strategy surpasses greedy selection.} We next ablate the Thompson sampler by replacing it with a greedy strategy that selects samples based on the mean estimate $\hat{p_i} = \alpha_i / (\alpha_i + \beta_i)$. The greedy approach overly exploits current knowledge and lacks exploration. In contrast, Thompson sampling naturally balances exploration and exploitation by stochastically sampling from the posterior to periodically select uncertain yet potentially valuable samples. Experimental results show that the greedy variant underperforms our Thompson sampling strategy by 1.50\% on average, underscoring the importance of maintaining exploration and exploitation balance for effective long-term learning.\\
\textbf{Ours two-scale prior decay exceeds last-update strategy.} Finally, we evaluate the two-scale prior decay mechanism against a "last-update" baseline that discards all historical data and uses only the most recent rollout to update $(\alpha_i, \beta_i)$. By contrast, our method retains historical information with time-aware decay, allowing it to robustly track sample difficulty in non-stationary environments. The performance gain of 1.65\% demonstrates the superiority of our proposed update strategy.

In summary, these ablation studies systematically validate the necessity of each component in VADE, confirming that the information gain objective, Thompson sampling, and two-scale prior decay collectively contribute to its strong performance.

\begin{table}[htbp]
  \caption{Ablation study on Qwen2.5VL-7B-Instruct with GRPO, evaluating the impact of key design choices on average performance across five benchmarks.}
  \label{tab:ablation_study}
  \centering
  \begin{tabular}{ll} 
    \toprule
    Method       & {AVG} \\  % 表头居中
    \midrule
    Ours &  \textbf{64.75} \\
    $p(1-p)$ & 63.66 (-1.09\%) \\
    greedy  & 63.25 (-1.50\%) \\
    last update & 63.10 (-1.65\%)\\
    \bottomrule
  \end{tabular}
\end{table}

\section{Conclusions}

%%%%%%%%%%%%%%%%%%%%%%%%%%%%%%%%%%
In this work, we identified and addressed the critical gradient vanishing problem in group-based RL, where identical rewards within groups cause advantage estimates to collapse, eliminating effective training signals. We proposed VADE, a variance-aware dynamic sampling framework via online sample-level difficulty estimation. By integrating online difficulty estimation using Beta distributions, Thompson sampling guided by an information-theoretic objective, and a two-scale prior decay mechanism, VADE proactively selects the most informative samples for each training batch without incurring extra rollout costs.

Extensive experiments on multimodal reasoning benchmarks show that VADE consistently outperforms strong baselines in both final performance and sample efficiency. Ablation studies validate the contribution of each component in our framework, confirming that the information gain objective, Thompson sampling, and two-scale prior decay collectively contribute to its strong performance. 

Notably, VADE can serves as a plug-and-play module compatible with existing group-based RL pipelines. For example, it can replace the dynamic sampling module in DAPO while preserving other configurations such as token-level policy optimization or clip-higher mechanisms. Future work will extend VADE to broader applications including mathematical reasoning and code generation, and explore integration with other policy optimization frameworks.

\bibliographystyle{unsrt}
\bibliography{ref}

\clearpage

%%%%%%%%%%%%%%%%%%%%%%%%%%%%%%%%%%%%%%%%%%%%%%%%%%%%%%%%%%%%

\appendix

\newpage
\appendix
\appendixpage
\startcontents[sections]
\printcontents[sections]{l}{1}{\setcounter{tocdepth}{2}}
\newpage
% \clearpage
% \setcounter{page}{1}
% \maketitlesupplementary

\section{Dataset Details}

We now detail the construction of our training dataset. We select 9 tasks from LLaVA-OneVision-Data\cite{llavaov} and incorporate the training sets of ChartQA\cite{masry2022chartqa} and ScienceQA\cite{ScienceQA}, ensuring no overlap between the training data and the evaluation benchmarks. The constructed dataset spans three distinct domains: mathematics, chart understanding, and scientific reasoning, comprising a total of 9,471 samples. The detailed data composition is presented in \Cref{tab:dataset_details}. All entries except ChartQA and ScienceQA correspond to subset names from LLaVA-OneVision-Data.

\begin{table*}[htbp]
  \caption{Composition of the constructed training dataset. All entries except ChartQA and ScienceQA correspond to subset from LLaVA-OneVision-Data. The dataset spans mathematical, chart-based, and scientific reasoning tasks.}
  \label{tab:dataset_details}
  \centering
  \begin{tabularx}{\linewidth}{l>{\centering\arraybackslash}p{3cm}X>{\centering\arraybackslash}p{3cm}}
    \toprule
    Domain & \#Total Samples & Tasks & \#Samples Per Task \\
    \midrule
    Math  & 5670 & 
    \begin{tabular}[t]{@{}l@{}}
    geo3k \\ 
    geo170k(qa) \\
    CLEVR-Math(MathV360K) \\
    GEOS(MathV360K) \\
    GeoQA+(MathV360K) \\
    Geometry3K(MathV360K)
    \end{tabular} &
    \begin{tabular}[t]{@{}c@{}}
    800 \\ 
    1,778 \\
    321 \\
    498 \\
    1,923 \\
    350
    \end{tabular} \\
    \midrule
    Chart & 2476 & 
    \begin{tabular}[t]{@{}l@{}}
    figureqa(cauldron,llava\_format) \\
    ChartQA\cite{masry2022chartqa} \\
    tabmwp(cauldron)
    \end{tabular} &
    \begin{tabular}[t]{@{}c@{}}
    957 \\
    1,300 \\
    219
    \end{tabular} \\
    \midrule
    Science & 1325 & 
    \begin{tabular}[t]{@{}l@{}}
    ScienceQA\cite{ScienceQA} \\
    ai2d(cauldron,llava\_format)
    \end{tabular} &
    \begin{tabular}[t]{@{}c@{}}
    1,100 \\
    225
    \end{tabular} \\
    \bottomrule
  \end{tabularx}
\end{table*}

\section{Implementation Details}

\Cref{tab:training_details} specifies the hyperparameter configuration used in our experiments. All training procedures employed identical parameter settings. All models were trained on NVIDIA H200 140GB GPUs.

\begin{table}[H]
  \caption{Hyperparameters for training Qwen2.5VL-7B/3B-Instruct models.}
  \label{tab:training_details}
  \centering
  \begin{tabular}{ll} 
    \toprule
    Hyperparameter  & Value \\  % 表头居中
    \midrule
     Train batch size & 512\\
     Max prompt length & 8192\\
     Max response length & 2048 \\
     Filter overlong prompts & True \\
     Rollouts per prompt & 8\\
     Total epoches & 15 \\
     Learning rate& 1e-6\\
     ppo\_mini\_batch\_size & 128 \\
     ppo\_micro\_batch\_size\_per\_gpu & 16 \\
     kl\_loss\_coef& 0.01\\
     tensor\_model\_parallel\_size& 2\\
     Rollout engine& vLLM\\
     GPU & NVIDIA H200 140G \\
     Train machines numbers & 1 \\
     GPU numbers per machine & 4\\
    \bottomrule
  \end{tabular}
\end{table}

\section{More Experimental Results}

\subsection{More Training Dynamics}

This section presents additional training dynamics results. 

\begin{figure*}[htbp]
   \centering
   \includegraphics[width=\linewidth]{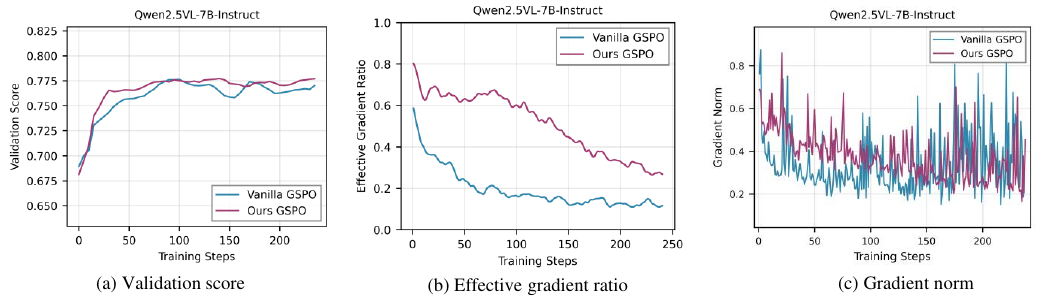}
   \caption{Training dynamics of Qwen2.5VL-7B-Instruct trained with the GSPO algorithm. (a) Validation score, illustrating convergence speed and final performance; (b) Effective gradient ratio, representing the proportion of data with non-uniform rewards (neither all-zero nor all-one) in each training batch, reflecting data efficiency; (c) Actor gradient norm, indicating the magnitude of gradient signals throughout training.}\label{fig:training_dynamic_ours_vanilla_7b_gspo}
\end{figure*}

\begin{figure*}[htbp]
   \centering
   \includegraphics[width=\linewidth]{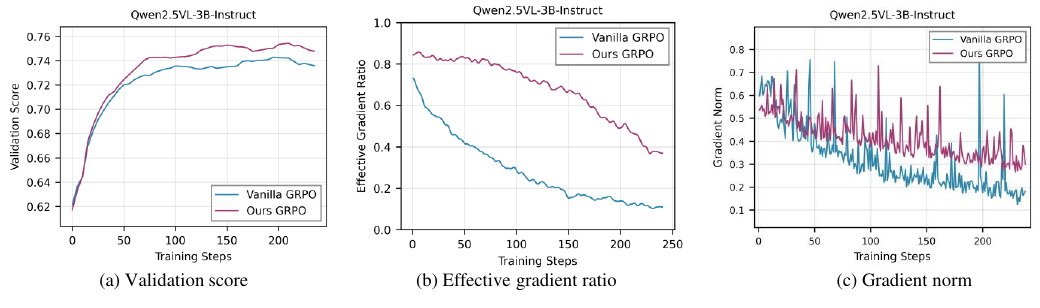}

   \caption{Training dynamics of Qwen2.5VL-3B-Instruct trained with the GRPO algorithm. (a) Validation score, illustrating convergence speed and final performance; (b) Effective gradient ratio, representing the proportion of data with non-uniform rewards (neither all-zero nor all-one) in each training batch, reflecting data efficiency; (c) Actor gradient norm, indicating the magnitude of gradient signals throughout training.}\label{fig:training_dynamic_ours_vanilla_3b_grpo}
\end{figure*}

\begin{figure*}[htbp]
   \centering
   \includegraphics[width=\linewidth]{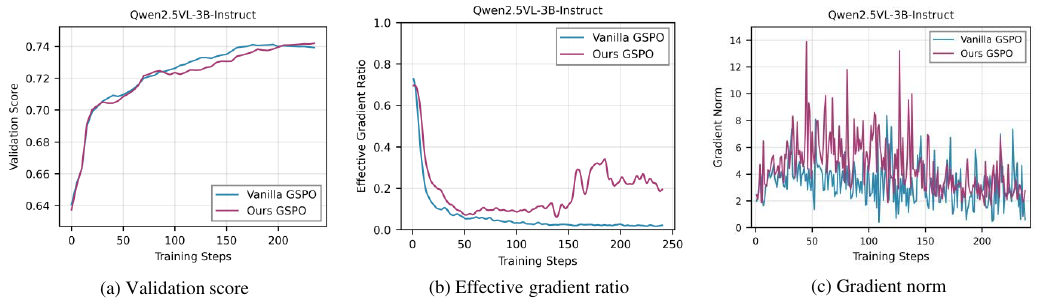}

   \caption{Training dynamics of Qwen2.5VL-3B-Instruct trained with the GSPO algorithm. (a) Validation score, illustrating convergence speed and final performance; (b) Effective gradient ratio, representing the proportion of data with non-uniform rewards (neither all-zero nor all-one) in each training batch, reflecting data efficiency; (c) Actor gradient norm, indicating the magnitude of gradient signals throughout training.}\label{fig:training_dynamic_ours_vanilla_3b_gspo}
\end{figure*}

\Cref{fig:training_dynamic_ours_vanilla_7b_gspo} shows the validation score, effective gradient ratio, and actor gradient norm curves comparing our method with vanilla method on Qwen2.5VL-7B-Instruct trained with GSPO. Similarly, \Cref{fig:training_dynamic_ours_vanilla_3b_grpo} presents corresponding results on Qwen2.5VL-3B-Instruct with GRPO, while \Cref{fig:training_dynamic_ours_vanilla_3b_gspo} shows results on Qwen2.5VL-3B-Instruct with GSPO. These results demonstrate that VADE consistently outperforms vanilla baselines across different model scales and algorithms, confirming its effectiveness and general applicability.

\begin{figure*}[htbp]
   \centering
   \includegraphics[width=\linewidth]{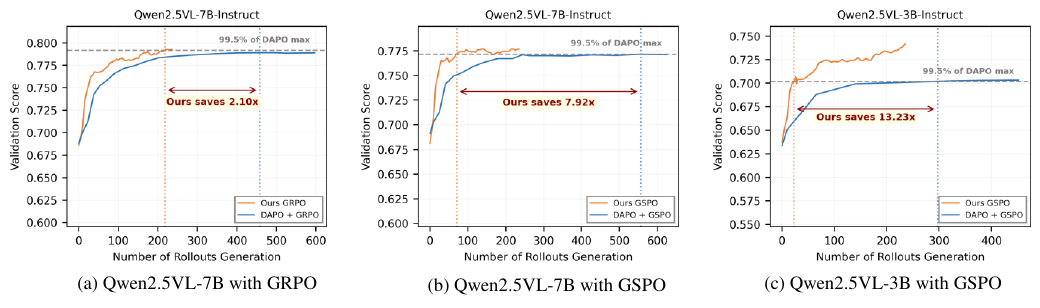}

   \caption{Extended comparison between VADE and DAPO across multiple configurations: (a) Qwen2.5VL-7B-Instruct with GRPO; (b) Qwen2.5VL-7B-Instruct with GSPO; (c) Qwen2.5VL-3B-Instruct with GSPO. VADE achieves competitive performance with significantly fewer rollout generations, demonstrating consistent efficiency gains across model scales and algorithms.}
   \label{fig:training_dynamic_ours_dapo_more}
\end{figure*}

Furthermore, \Cref{fig:training_dynamic_ours_dapo_more} extends the computational efficiency comparison between VADE and DAPO across multiple configurations: Qwen2.5VL-7B-Instruct with GRPO, Qwen2.5VL-7B-Instruct with GSPO, and Qwen2.5VL-3B-Instruct with GSPO. Our method consistently achieves superior performance with significantly fewer rollout generations under all these settings, highlighting its substantial advantage in training efficiency and general applicability.

\subsection{Detailed Ablation Study Results}

This section provides complete ablation results across all benchmarks in \Cref{tab:ablation_study_details}. Following the same format as \Cref{table:main_results}, we report average@4 scores for each benchmark. Our full configuration consistently outperforms all ablation variants, demonstrating the effectiveness of each component in VADE's design.

\begin{table*}
  \caption{Detailed ablation study results on Qwen2.5VL-7B-Instruct with GRPO, reporting average@4 performance across all evaluation benchmarks. Highlighted cells indicate whether each variant \colorbox{green!10}{improves} or \colorbox{red!5}{degrades} performance compared to our full VADE design.}
  \label{tab:ablation_study_details}
  \centering
  \begin{tabular}{lcccccc}
    \toprule
    Method & MathVista & MathVerse & MathVision & ScienceQA & ChartQA & AVG \\
    \midrule
    Ours & 75.10 & 45.65 & 25.79 & 91.67 & 85.56 & 64.75 \\
    $p(1-p)$ & \colorbox{red!5}{73.07} & \colorbox{red!5}{45.57} & \colorbox{red!5}{24.37} & \colorbox{red!5}{90.38} & \colorbox{red!5}{84.92} & \colorbox{red!5}{63.66} \\
    greedy & \colorbox{red!5}{73.15} & \colorbox{red!5}{44.72} & \colorbox{red!5}{24.74} & \colorbox{red!5}{90.32} & \colorbox{red!5}{84.30} & \colorbox{red!5}{63.25} \\
    last update & \colorbox{red!5}{71.33} & \colorbox{red!5}{45.11} & \colorbox{red!5}{22.25} & \colorbox{red!5}{91.22} & \colorbox{green!10}{85.60} & \colorbox{red!5}{63.10} \\
    \bottomrule
  \end{tabular}
\end{table*}

% 73.15 	44.72 	23.74 	90.32 	84.30 	63.25 

% greedy & 73.15 & 45.82 & 24.94 & 91.72 & 85.80 & 64.29 \\ 

% \section{Code Availability}
% The complete implementation of our VADE framework, along with detailed instructions and configurations for reproducing all experimental results, is publicly available at:
% https://anonymous.4open.science/r/VADE-3F7B/README.md

\end{document}